\documentclass{article} 
\usepackage{nips10submit_e,times}

\usepackage{amsmath,amssymb}
\usepackage{color}
\usepackage{mydef,times}
\usepackage{subfigure}

\def \fbf{ \mathbf{f} }
\def \tbf{ \mathbf{t} }
\def \GP{ \mathcal{G} \mathcal{P} }
\def \Mbf{ \mathbf{M}}
\def \Kti{ \widetilde{\Kbf}}

\def\nubf{\boldsymbol{\nu}}

\title{Multiple Gaussian Process Models}

\author{
Cedric Archambeau\\
Xerox Research Centre Europe\\
6, Ch. de Maupertuis, 38240 Meylan, France\\
\texttt{cedric.archambeau@xerox.com} \\
\And
Francis Bach \\
       INRIA-WILLOW Project-Team \\
       23, Av. d'Italie, 75214 Paris, France\\
\texttt{francis.bach@inria.fr}
}

\nipsfinalcopy 

\begin{document}

\maketitle

\begin{abstract}
We consider a Gaussian process formulation of the multiple kernel learning problem. The goal is to select the convex combination of kernel matrices  that best explains the data and by doing so improve the generalisation on unseen data. Sparsity in the kernel weights is obtained by adopting a hierarchical Bayesian approach: Gaussian process priors are imposed over the latent functions and generalised inverse Gaussians on their associated weights. This construction is equivalent to imposing a product of heavy-tailed process priors over function space. A variational inference algorithm is derived for regression and binary classification.
\end{abstract}

\section{Introduction}

Kernel-based methods are well-established tools for supervised learning, allowing to perform various tasks, such as regression or binary classification, with linear and non-linear predictors. Like most statistical models, kernel-based methods can be considered within two frameworks: in the frequentist approach, estimators are obtained by minimizing a regularized empirical risk, leading e.g. to kernel ridge regression or the support vector machine \cite{Shawe-Taylor04,Scholkopf00}; in the Bayesian approach, Gaussian processes (GPs) provide a Bayesian interpretation to kernel-based methods~\cite{Rasmussen06}, with the potential to learn the kernel parameters from the data without having to use cross-validation.

Crucial to the predictive performance of kernel methods is the choice of the kernel function. 
In the Bayesian setting, the kernel function (often called covariance function) determines the correlations between the predictions we make. Assuming that the predictor's smoothness is fully specified by these correlations can be formalised by a Gaussian process imposed over function space. Techniques based on automatic relevance determination have been successful at learning the parameters of kernel functions such as the individual length scales of the squarred exponential kernel~\cite{Rasmussen06}. In the frequentist setting, a specific parameterization of kernel functions has led to a significant amount of work, namely positive linear combination of pre-defined kernel functions (or kernel matrices), leading to the multiple kernel learning (MKL) framework~\cite{Lanckriet:JMLR04,Bach:ICML04}. The first contribution of this paper is to propose a Gaussian process (GP) formulation of the multiple kernel learning framework, which we refer to as \emph{multiple Gaussian process} (MGP) models. Its second contribution is to provide a framework to consider all $\ell_p$-norms at once and to determine \emph{from data} whether we should use a sparsity-inducing prior or not. Currently, there is no consensus in the frequentist community on how to choose the type of regularization. In practice, however, the choice of the regulariser leads to solutions of very different kinds. For example, when considering an $\ell_1$-norm a sparse solution will be obtained whether or not it is supported by the data. Obviously, if all kernels are important for prediction, this will be detrimental~\cite{Gehler:ICCV09}.

\section{Multiple Gaussian process model for regression} \label{sect: MGP regression}

Let $\{y_n\}_{n=1}^N$ be the set of noisy targets and $\{\xbf_{n1},\ldots,\xbf_{nP}\}_{n=1}^N$ the set of features, which are assumed to be non-random column vectors. We consider a weighted linear model of the $P$ feature vectors with i.i.d.~Gaussian noise:
\begin{align}
\ybf|\Xbf_1,\ldots,\Xbf_P,\wbf_1,\ldots,\wbf_P
	&\sim \Ncal({\textstyle\sum_p}\Xbf_p\wbf_p,\tau^{-1}\Ibf_N) ,\label{eq: like lin}
\end{align}
where $\ybf=(y_1,\ldots,y_N)^\top$ and $\Ibf_N$ is the identity matrix of dimension $N$. The weights associated to the feature matrix $\Xbf_p\in\mathbb{R}^{N\times D_p}$ are denoted by $\wbf_p\in\mathbb{R}^{D_p}$ and the residual precision by $\tau$.

The case of interest is the one where the weight vectors are sparse, i.e., many of their elements are (nearly) zero. However, we do not know a priori the degree of sparsity. From a Bayesian perspective, the spike and slab prior is the golden standard for inducing sparsity. Here, follow a different approach and choose the prior $p(\wbf_p)$ to be a Gaussian scale mixture~\cite{Andrews:JRSS74} centred at zero. In effect we approximate the spike and slab prior by a continuous prior which favours sparse solutions. Although zero probability mass is put on exact zero values, the use of heavy-tailed priors allows us to infer large, as well as quasi zero values for the kernel weights.

Formally, we impose the following product of zero-mean Gaussian scale mixtures on the weights:
\begin{align}
\wbf|\gambf
	&\sim{\textstyle\prod_p}\Ncal(\mathbf{0},\gamma_p^{-1}\Ibf_{D_p}) ,
	&\gambf
	&\sim{\textstyle\prod_p}\Ncal^{-1}(\omega,\chi,\phi) ,\label{eq: gam prior}
\end{align}
where $\gambf=(\gamma_1,\ldots,\gamma_P)^\top$ is the vector of unobserved scale variables on which independent generalised inverse Gaussian densities (see Appendix~\ref{sect: inverse Gaussian}) are imposed. The marginal $p(\wbf)$ is then a symmetric generalised hyperbolic density \cite{Hu05}, which has fat tails compared to the Gaussian. This family contains the Student-$t$, the Laplace, the Gamma-variance and Jeffrey's as special cases.

Given this probabilistic model, one can integrate out $\wbf$, leading to a closed form expression for the marginal density of the observations:
\begin{align}
\ybf| \gambf
&\sim \Ncal\big(\mathbf{0},{\textstyle\sum_{p}} \gamma_p^{-1}\Kbf_p + \tau^{-1}\Ibf_N\big) ,\label{eq: marginal likelihood}
\end{align}
where $\Kbf_p = \Xbf_p \Xbf_p^\top \in \mathbb{R}^{N \times N}$ is the kernel matrix associated with the $p$-th feature matrix. Clearly, the marginal density is well-defined for any set of valid kernel matrices since $\gamma_p>0$ for all $p$.

The linear additive model defined in (\ref{eq: like lin}) corresponds to the \emph{weight-space} view representation of the multiple Gaussian process model (MGP). From (\ref{eq: marginal likelihood}), however, we see that the marginal density only depends on the linear kernel matrices $\{\Kbf_p\}_{p=1}^P$. Hence, the model can be generalised to a non-linear additive model by replacing these linear kernel matrices by non-linear ones. This new representation corresponds to the multiple \emph{function-space} view representation.

Let $\{f_p(\cdot)\}_{p=1}^P$ be a set of $P$ latent functions on which we impose scaled Gaussian process priors:
\begin{align}
f_p(\cdot)|\gamma_p\sim\GP(0,\gamma_p^{-1}k_p(\cdot,\cdot)) ,
\end{align}
for all $p$. The functions $\{k_p(\cdot,\cdot)\}_{p=1}^P$ are covariance functions, which are also valid kernel functions \cite{Rasmussen06}. Again we consider i.i.d.~Gaussian noise, but assume $\{y_n\}_{n=1}^N$ are noisy observations of a sum of $P$ latent \emph{function} values $\fbf_p\in\mathbb{R}^N$. The likelihood function and the MGP prior are given by
\begin{align}
\ybf|\fbf,\tau
&\sim\Ncal({\textstyle\sum_p} \fbf_p,\tau^{-1}\Ibf_N)=\Ncal(\Mbf\fbf,\tau^{-1}\Ibf_N) ,\label{eq: likelihood}\\
\fbf|\gambf
&\sim{\textstyle\prod_p}\Ncal(\mathbf{0},\gamma_p^{-1}\Kbf_p )=\Ncal(\mathbf{0},\Kti) ,\label{eq: GP prior}
\end{align} 
where $\fbf^\top=(\fbf_1^\top,\ldots,\fbf_P^\top)$, $\Mbf=(\mathbf{1}_P^\top\otimes\Ibf_N)$ and $\Kti=\diag\{\gamma_1^{-1}\Kbf_1,\ldots,\gamma_P^{-1}\Kbf_P\}$. Vector $\mathbf{1}_P$ is the unit vector of dimension $P$ and the operator $\otimes$ denotes the Kronecker product. The prior on $\gambf$ is still given by (\ref{eq: gam prior}) and the marginal $p(\ybf|\gambf)$ has the same form as in the weight-space view representation.

The MGP model corresponds to imposing $P$ independent \emph{non-Gaussian} process priors over function space. If we condition on the corresponding scale variable, any finite subset of latent function values is distributed according to a multivariate Gaussian marginal. For any of these marginals one can integrate out the scale variable, such that any finite set of latent function values is distributed according  to a product of $P$ independent multivariate Gaussian scale mixture densities:
\begin{align}
\fbf	&\sim\prod_p\int \Ncal(\mathbf{0},\gamma_p^{-1}\Kbf_p)\ p(\gamma_p)\ d\gamma_p
	\propto\prod_p{\textstyle\frac{K_{\omega+\frac{N}{2}}\left(\sqrt{\chi(\phi+\fbf_p^\top\Kbf_p^{-1}\fbf_p)}\right)}
	{\left(\sqrt{(\phi+\fbf_p^\top\Kbf_p^{-1}\fbf_p)/\chi}\right)^{\omega+\frac{N}{2}}}} .\label{eq: effective prior}
\end{align}
where $K_{\omega}(\cdot)$ is the modified Bessel function of the second kind. Hence, the prior measure imposed over function space is a heavy-tailed one, known as the generalised hyperbolic measure~\cite{Barndorff:PRS77,Hu05}. The Gaussian process is recovered for $\omega\rightarrow\infty$ and the symmetric multivariate zero-mean hyperbolic process is obtained for $\omega=-1$. Other special cases include the multivariate Gamma-variance process ($\omega<0$ and $\phi=0$), the multivariate Laplace process ($\omega=-1$ and $\phi=0$), the multivariate Student-$t$ process ($\omega>0$ and $\chi=0$) and the multivariate Cauchy process ($\omega=1/2$ and $\chi=0$).

The most straightforward approach for the estimation of $\gambf$ is to use type II maximum a postriori (or type II maximum likelihood in absence of prior on $\gambf$ as adopted in \cite{Kapoor:IJCV10}). The optimisation can be performed using standard nonlinear optimisation tools, but the regulariser needs to be chosen in advance. Instead, we turn our attention to the inference problem of these parameters from data.  We view $\gambf$ as a latent variable  and the desired level of sparsity is learnt from the data by optimising the hyperparameters by type II maximum likelihood (ML).

\section{Variational inference with type II maximum likelihood}\label{sect: inference}

We follow a mean field approach \cite{Beal03,Bishop06}. In order to find an analytically tractable solution, the posterior over the latent function values $\fbf$ and the scale vector $\gambf$ is assumed to factorise given the data, that is $q(\fbf,\gambf)= q(\fbf)\prod_pq(\gamma_p)$. It can be shown that the variational posteriors maximising the negative variational free energy (a lower bound to the log-marginal likelihood) are given by $q(\fbf)=\Ncal(\mubf,\Sigbf)$ and $q(\gambf)=\prod_p\Ncal^{-}(\omega_p,\chi_p,\phi_p)$. The parameters are defined as
\begin{align*}
\mubf
	&=\tau\Sigbf\Mbf^\top\ybf ,
&\Sigbf
	&=\big(\bar{\Kbf}^{-1}+\tau\Mbf^\top\Mbf\big)^{-1} ,
&\omega_p
	&=\omega+N/2 ,
&\chi_p
	&=\chi ,
&\phi_p
	&=\phi+\langle\fbf_p^\top\Kbf_p^{-1}\fbf_p\rangle ,
\end{align*}
where $\bar{\Kbf}=\diag\{\langle\gamma_1\rangle^{-1}\Kbf_1,\ldots,\langle\gamma_P\rangle^{-1}\Kbf_P\}$.

The predictive MGP is obtained by assuming that $q(\gambf)$ is peaked around its mean such that $q(f(\xbf))\approx q(f(\xbf)|\langle\gambf\rangle)$. The true predictive density can then be approximated by the following analytically tractable integral:
\begin{align}
y(\xbf)|\ybf
	&\sim\int p(y(\xbf)|f(\xbf))\ q(f(\xbf))\ df(\xbf)
	=\GP\big(m(\xbf),v(\xbf)+\tau^{-1}\big) .\label{eq: predictive GP}
\end{align}
where
\begin{align}
m(\xbf)
	&= {\textstyle\sum_p}\langle\gamma_p\rangle^{-1}\kbf_p(\xbf_p)^\top \Bbf^{-1}\ybf  ,\label{eq: pred mean}\\
v(\xbf)
	&= {\textstyle\sum_p}\langle\gamma_p\rangle^{-1}k_p(\xbf_p,\xbf_p)
	- {\textstyle\sum_p\sum_q}\langle\gamma_p\rangle^{-1}\langle\gamma_q\rangle^{-1}\kbf_p(\xbf_p)^\top \Bbf^{-1}\kbf_q(\xbf_q)\big\} ,\label{eq: pred var}
\end{align}
with $\Bbf=\sum_r\langle\gamma_r\rangle^{-1}\Kbf_r + \tau^{-1}\Ibf_N$. From these expression we see that the posterior mean and variance have the same form as in standard GP regression; the kernel is simply replaced by a convex combination of kernels. Note, moreover, that the expression $m(\xbf)$ has the same form as the one we would obtain with a frequentist method such as  kernel ridge regression.

The ML II updates for the hyperparameters are obtained by solving the following expressions (which are simple root finding equations, with unique solutions, hence easily solved by binary search):
\begin{align}
\!\! \omega:
&\ \textstyle P\ln\sqrt{\frac{\phi}{\chi}}\! -\! P\frac{d\ln K_{\omega}(\sqrt{\chi\phi})}{d\omega}
\!+\!\sum_p\langle\ln\gamma_p\rangle =0 ,\label{eq: M step omega}\\
\chi:
&\ \textstyle \frac{P\omega}{\chi}
-\frac{P}{2}\sqrt{\frac{\phi}{\chi}}R_{\omega}(\sqrt{\chi\phi})
+\frac{1}{2}\sum_p\langle\gamma_p^{-1}\rangle =0 ,\label{eq: M step chi}\\
\phi:
&\ \textstyle -\frac{P}{2}\sqrt{\frac{\chi}{\phi}}R_{\omega}(\sqrt{\chi\phi})
+ \frac{1}{2}\sum_p \langle\gamma_p\rangle =0 ,\label{eq: M step phi}
\end{align}
where $R_\omega(\cdot)=K_{\omega+1}(\cdot)/K_\omega(\cdot)$. These updates are obtained by direct maximising of the variational bound. The update for $\tau$ is obtained in the same manner.

\section{MGP for classification}

We restrict ourselves to binary classification and consider a scaled probit model in which the likelihood is derived from the Gaussian cumulative density. A probit model is equivalent to a Gaussian noise and a step function likelihood \cite{Albers:JASA93,Opper:NC00}.

Let $\{t_n\}_{n=1}^N$ be the class labels, with $t_n\in\{-1,+1\}$ for all $n$. The likelihood (\ref{eq: likelihood}) is replaced by 
\begin{align}
\tbf|\ybf
&\sim\textstyle\prod_{n} I(t_ny_n) ,
&\ybf|\fbf
&\sim\Ncal({\textstyle\sum_p}\fbf_p,\tau^{-1}\Ibf_N) ,
\end{align}
where $I(z)=1$ for $z\geqslant 0$ and $0$ otherwise.

As in the case of regression, we consider a mean field approximation and assumes the posterior is of the form $q(\ybf)q(\fbf)q(\gambf)$. We further assume the variational posterior $q(\ybf)$ is a product of truncated Gaussians (see Appendix~\ref{sect: truncated Gaussian}):
\begin{align}
q(\ybf)
	&\propto\textstyle \prod_n I( t_ny_n) \Ncal( \nu_n, \lambda_n )
	= \left(\prod_{t_n=+1}\Ncal_+( \nu_n, \lambda_n )\right)\left(\prod_{t_n=-1}\Ncal_-( \nu_n, \lambda_n )\right)   ,
\end{align}
where $\nu_n=\sum_p\langle f_p(\xbf_{np})\rangle$ and $\lambda_n=1/\tau$. The posterior mean and the posterior covariance of $\fbf$ are unchanged, except that $\ybf$ is replaced by $\nubf_{\pm}$. The elements of $\nubf_{\pm}$ are defined in (\ref{eq: trun mean}). The posterior $q(\gambf)$ and the updates for the hyperparameters are identical to the ones in MGP regression.

In Bayesian classification the label with highest probability $P(t(\xbf)|\tbf)$ is selected. Since an exact computation is analytically intractable, we assume the posteriors $q(\ybf)$ and $q(\gambf)$ are highly peaked around their mean leading to the following classification rule:
\begin{align}
P(t(\xbf)=\pm1|\tbf)
&\approx \textstyle P(t(\xbf)=\pm1|\tbf,\nubf_{\pm},\langle\gambf\rangle) = \Phi\big(\pm m(\xbf)/\sqrt{v(\xbf)+\tau^{-1}}\big) ,
\end{align}
where $m(\xbf)$ and $v(\xbf)$ are as before with $\ybf$ replaced by $\nubf_{\pm}$. Deciding whether the label is $-1$ or $+1$ is equivalent to using the sign of $m(\xbf)$ as the decision rule.

\section{Discussion}

We compare frequentist and Bayesian approaches to kernel combination. We demonstrate the flexibility and the performance of the MGP models on the following two detat sets:

\textbf{Toy regression data.} We generate random functions from the Hilbert spaces induced by 10 Laplacian kernels and add Gaussian i.i.d.~noise; we show results on three different settings: a sparse problem, where only one kernel is used to generate the response, a semi-sparse problem with 3 functions are used and a non-sparse problem where all ten functions are active. Table~\ref{table: regression} compares several MGP models and several cross-validated MKL models with fixed regularisation norms. Fig.~\ref{fig: regression student} in the Appendix shows that the hierarchical Bayesian approach is able to adapt to the sparsity of the data. 

\begin{table}[t]
\caption{Average root mean square error for toy regression data (lower is better). The multiple Laplace process performs worse than the Student-$t$ and the Gamma-variance process when the generating process is sparse. ARD performs poorly when the generating process is not sparse. In the case of MKL the prior choice of the regulariser leads to more sensitivity to model misspecifications.}
\label{table: regression}
\begin{center}
\begin{tabular}{lcccr}
Number of active kernels & 1 out of 10 & 3 out of 10 & 10 out of 10 \\
\\ \hline \\
Multiple Student-$t$        & .033 ($\pm .027$) & .067  ($\pm .032$) & .719  ($\pm .221$)\\
Multiple Laplace       & .034 ($\pm .028$) & .076  ($\pm .035$) & .704  ($\pm .204$)\\
Multple Gamma-variance        & .033 ($\pm .027$) & .067  ($\pm .032$) & .719  ($\pm .223$)\\
ARD       & .033 ($\pm .027$) & .066  ($\pm .031$) & .746  ($\pm .223$)\\
MKL $\ell_1$                    &  .037 ($\pm .032$) & .066 ($\pm .030$) & .720 ($\pm .203$) \\
MKL $\ell_2$                    &  .830 ($\pm .655$) &  .831 ($\pm .399$) & .762 ($\pm .251$) \\
MKL $\ell_{4/3}$             &  .097 ($\pm .062$) & .233 ($\pm .098$) & .719 ($\pm .238$) \\
\end{tabular}
\end{center}
\end{table}

\begin{figure}[t]
\vspace{-1.2cm}
\begin{center}
\subfigure[Typical roc curve (class 102).]{\includegraphics[scale=.5]{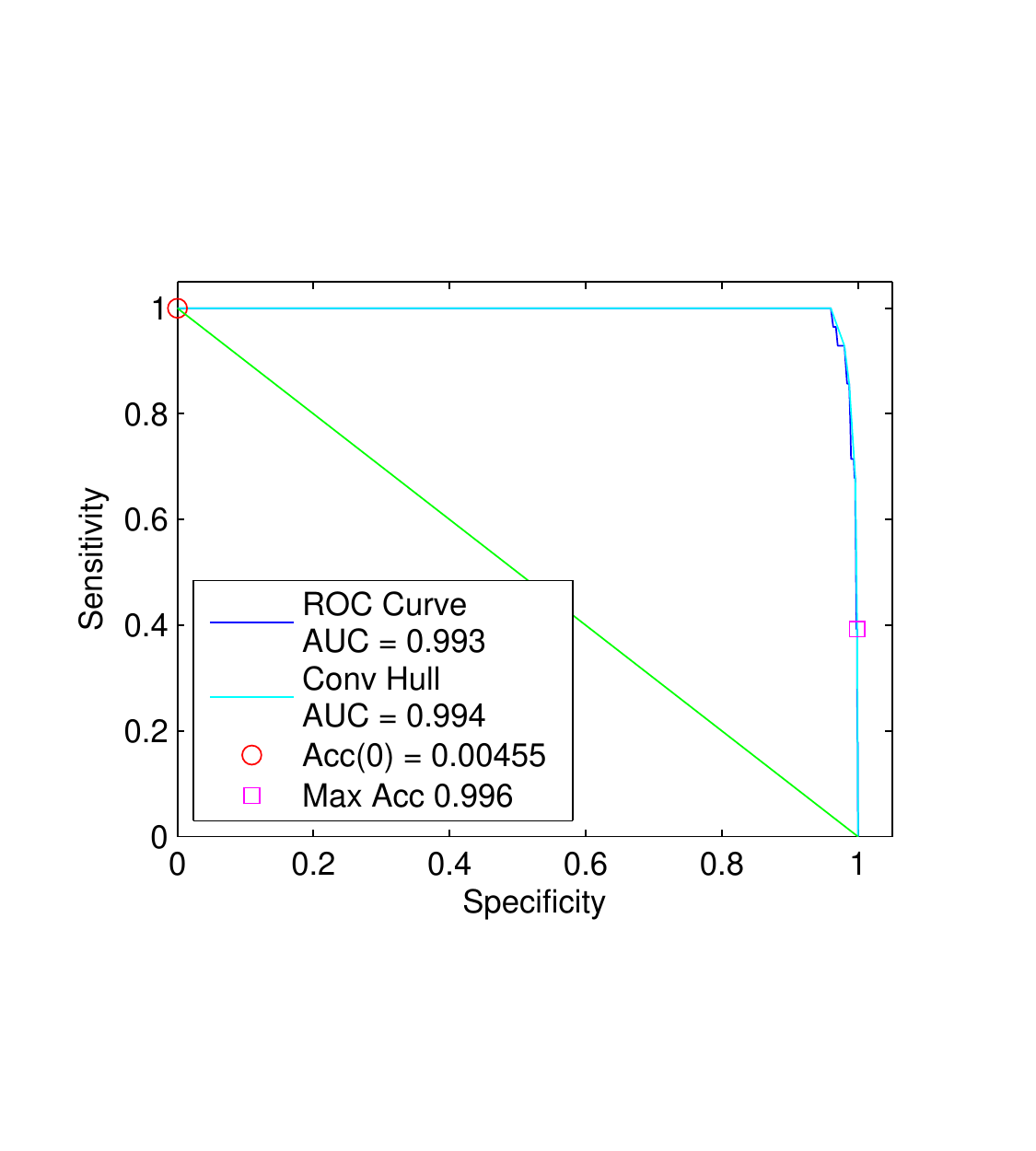}}
\subfigure[Worse roc curve (class 96).]{\includegraphics[scale=.5]{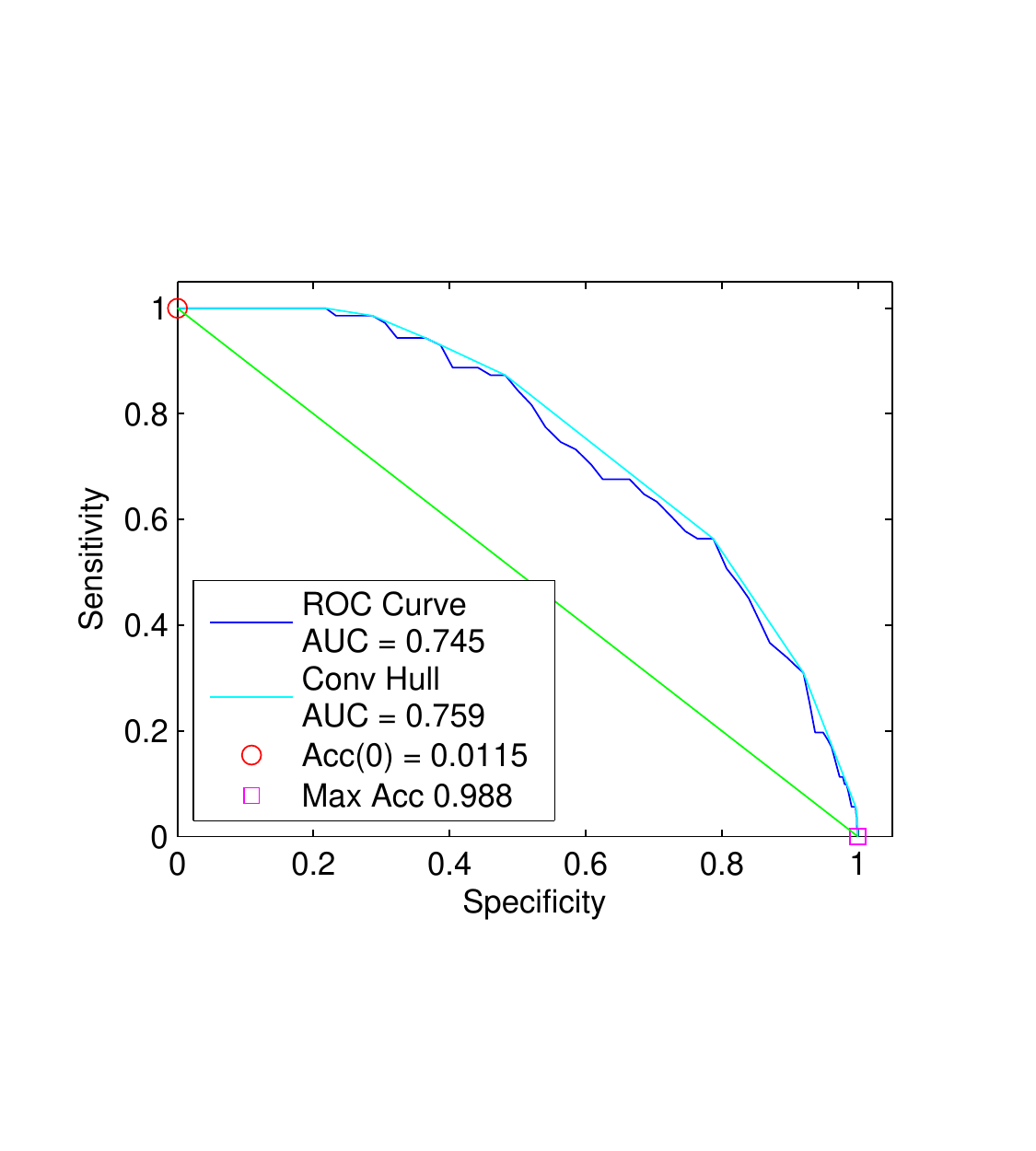}}
\end{center}
\caption{Flowers data. ROC curves obtained for Student-$t$ process one-versus-all classification for two flower classes. The ROC curves obtained for Gamma-variance process are slightly better.}
\label{fig: roc student}
\end{figure}

\textbf{Flowers data set.}\footnote{www.robots.ox.ac.uk/~vgg/data/flowers/102/} Due to a lack a space we do not describe the data and the features, but only mention it is a standard MKL benchmark for multi-class image classification. For each of the $102$ flower classes we learn a one-versus-all classifier. Fig.~\ref{fig: roc student} shows the ROC curve for two classes when considering a Student-$t$ process, for which we obtained an average $\mathrm{AUC}= .948\pm.057$. The average $\mathrm{AUC}$ for Gamma-variance process and ARD are respectively given by $= .957\pm.050$ and $.947\pm.058$. All are better than state-of-the-art MKL results \cite{Nilsback:ICCVGIP08}.

%

\small{
\bibliographystyle{alpha}
\bibliography{C:/Local/Archambeau/Papers/mybib}
}

\newpage
\appendix

\section{Generalised inverse Gaussian density}\label{sect: inverse Gaussian}

The generalised inverse Gaussian density is defined as follows \cite{Jorgensen82}:
\begin{align}
x\sim\Ncal^{-1}(\omega,\chi,\phi)
= \frac{\chi^{-\omega}(\sqrt{\chi\phi})^\omega}{2K_\omega(\sqrt{\chi\phi})}x^{\omega-1}
e^{-\frac{1}{2}(\chi x^{-1}+\phi x)} ,
\end{align}
where $x>0$ and $K_\omega(\cdot)$ is the modified Bessel function of the second kind with index $\omega\in\re$. Depending on the value taken by $\omega$, we have the following constraints on $\chi$ and $\phi$:
\begin{align*}
\left\{\begin{array}{ll}
\omega>0: & \chi\geqslant 0,\ \phi>0 ,\\
\omega=0: & \chi> 0,\ \phi>0 ,\\
\omega<0: & \chi> 0,\ \phi\geqslant 0 .
\end{array}{}\right.
\end{align*}
Let us define $R_{\omega}(\cdot)= K_{\omega+1}(\cdot)/K_{\omega}(\cdot)$. The following expectations are useful:
\begin{align}
\langle x\rangle
	&=\sqrt{\frac{\chi}{\phi}}R_{\omega}(\sqrt{\chi\phi}) ,
	&\langle x^{-1}\rangle
	&=\sqrt{\frac{\phi}{\chi}}R_{-\omega}(\sqrt{\chi\phi}) ,
	&\langle\ln x\rangle
	&= \ln\sqrt{\frac{\chi}{\phi}}+\frac{d\ln K_{\omega}(\sqrt{\chi\phi})}{d\omega} ,\label{eq: expectations GIG}
\end{align}
When $\chi=0$ and $\omega>0$, the generalised inverse Gaussian density reduces to the Gamma density. When $\phi=0$ and $\omega<0$, it reduces to the inverse Gamma density. The expectations simplify also.

\section{Truncated Gaussian density}\label{sect: truncated Gaussian}

The (positive/negative) truncated Gaussian density is defined as follows:
\begin{align}
\Ncal_{\pm}( \mu, \sigma^2)
&=  \Phi(\pm\mu/\sigma)^{-1} \Ncal( \mu, \sigma^2) ,
\end{align}
where $\Phi(a)=\int_{-\infty}^a\Ncal(0,1)dz$ is the cumulative density of the unit Gaussian.

Let $x_{\pm}  \sim \Ncal_{\pm}( \mu, \sigma^2)$. The mean and variance are given by
\begin{align}
\langle x_{\pm}\rangle
	&=  \mu \pm \sigma^2 \Ncal_{\pm}( 0 | \mu, \sigma^2) ,\label{eq: trun mean}\\
\langle( x_{\pm} - \langle x_{\pm}\rangle)^2\rangle
	&=  \sigma^2 \mp \sigma^2 \mu \Ncal_{\pm}( 0 | \mu, \sigma^2)
	- \sigma^4 \Ncal_{\pm}( 0 | \mu, \sigma^2)^2 .\label{eq: trun var}
\end{align}

\section{Example of the inferred kernel weights}

\begin{figure}[h]
\begin{center}
\subfigure[1 active kernel.]{\includegraphics[scale=.4]{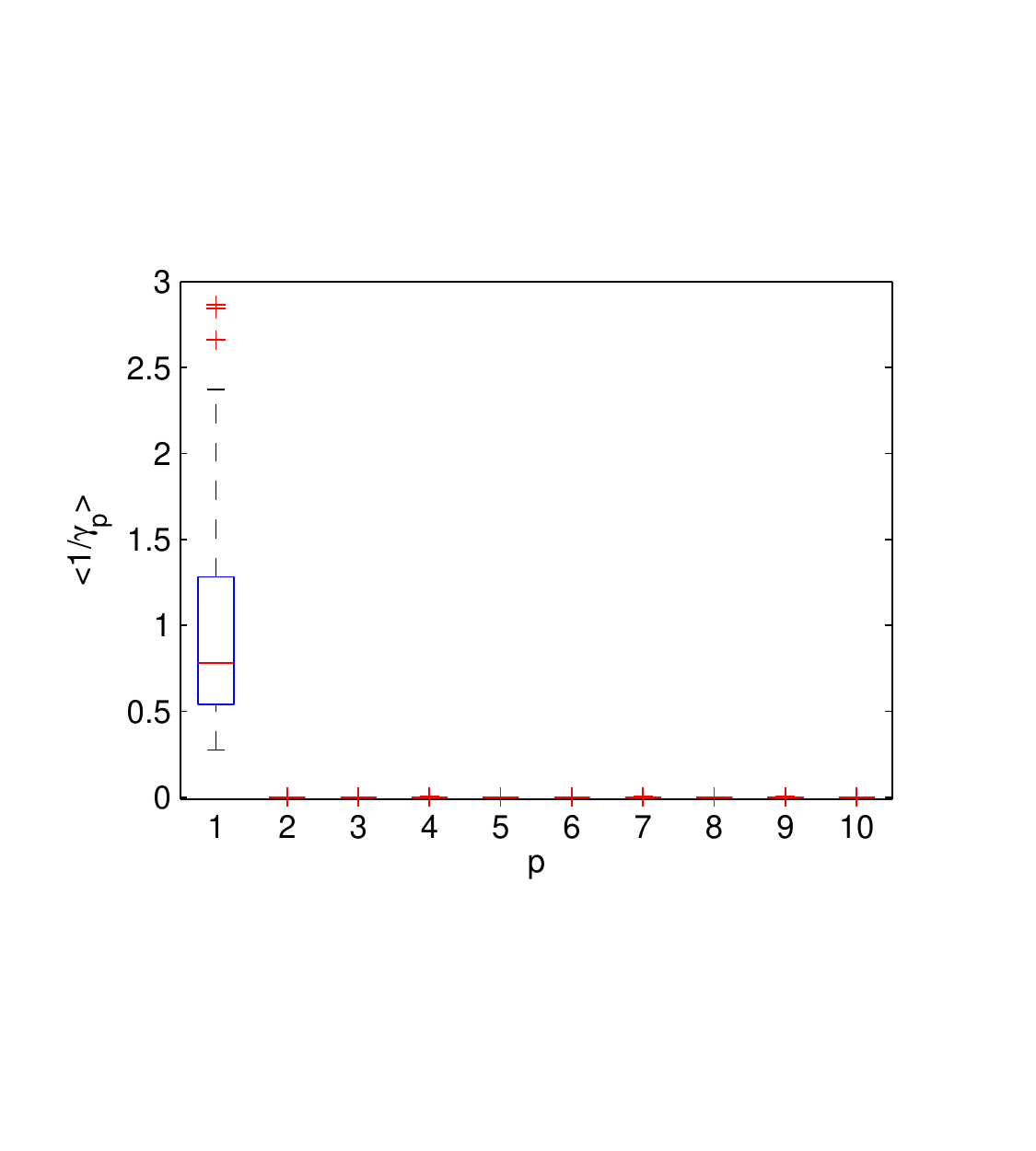}}
\subfigure[3 active kernels.]{\includegraphics[scale=.4]{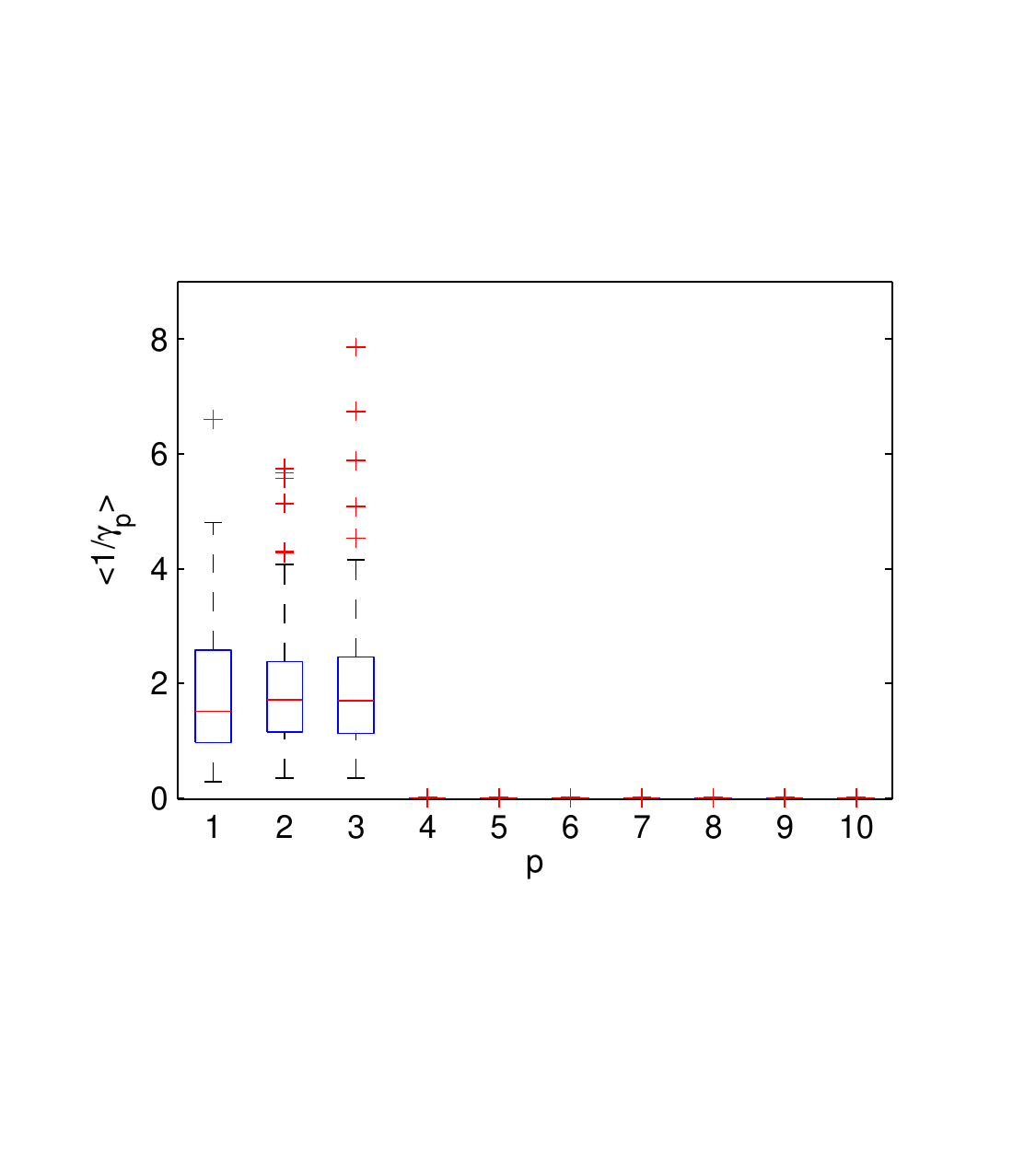}}
\subfigure[10 active kernels.]{\includegraphics[scale=.4]{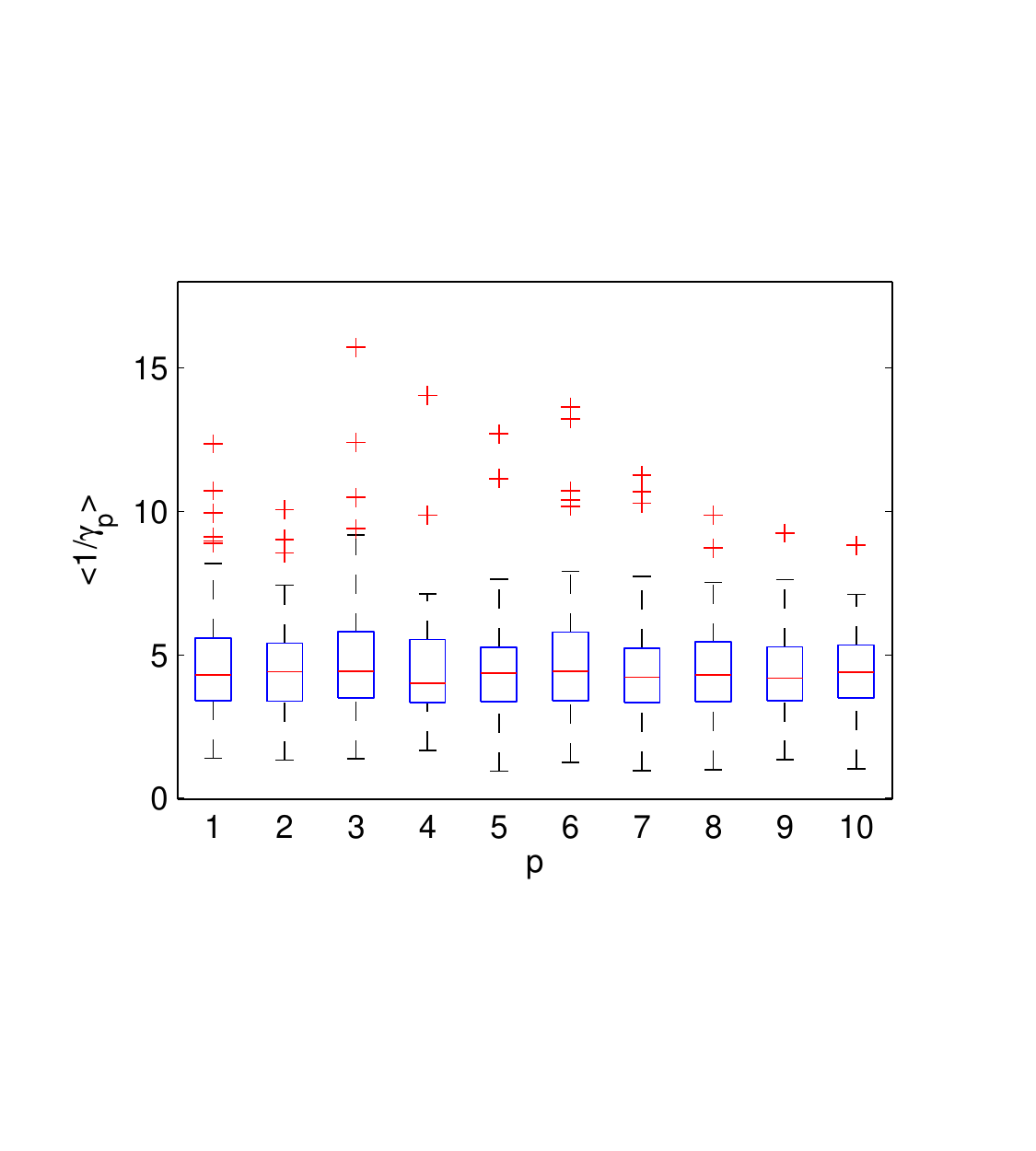}}
\end{center}
\caption{Toy regression data. Shown are the box-and-whisker plots of the expected weight for each kernel when consider a Gamma prior (Student-$t$ process). Other generalised inverse Gaussian priors perform as well. The MGP model was run on hundred different data set realisations.}
\label{fig: regression student}
\end{figure}

\end{document}